\documentclass[letterpaper, 10 pt, conference]{ieeeconf}  

\IEEEoverridecommandlockouts                              

\usepackage{graphicx} 
\usepackage{tikz}

\usepackage{amsmath,amsfonts,amssymb}

\title{\LARGE \bf
Online 3D Bin Packing Reinforcement Learning Solution with Buffer}

\author{Aaron Valero Puche$^{1}$ and Sukhan Lee$^{1*}$
\thanks{*This work is supported, in part, by "3D Bin Packing with Deep Reinforcement Learning" project funded by Hyundai Robotics Co. Ltd., in part, by “Edge Brain Based Intelligent Manufacturing” project IITP-2022-0-00067, in part, by AI Graduate School Program, Grant No.2019-0-00421, and by ICT Consilience Program, IITP-2020-0-01821, of the Institute of Information and Communication Technology Planning Evaluation (IITP), sponsored by the Korean Ministry of Science and Information Technology (MSIT).}
\thanks{$^{1}$ Authors from the Artificial Intelligence School, Sungkyunkwan University, Suwon, South Korea, * Corresponding author: Sukhan Lee 
{\tt\small Lsh1@skku.edu}}}

\begin{document}

\maketitle
\thispagestyle{empty}
\pagestyle{empty}

\begin{abstract}
The 3D Bin Packing Problem (3D-BPP) is one of the most demanded yet challenging problems in industry, where an agent must pack variable size items delivered in sequence into a finite bin with the aim to maximize the space utilization. It represents a strongly NP-Hard optimization problem such that no solution has been offered to date with high performance in space utilization. In this paper, we present a new reinforcement learning (RL) framework for a 3D-BPP solution for improving performance. First, a buffer is introduced to allow multi-item action selection. By increasing the degree of freedom in action selection, a more complex policy that results in better packing performance can be derived. Second, we propose an agnostic data augmentation strategy that exploits both bin item symmetries for improving sample efficiency. Third, we implement a model-based RL method adapted from the popular algorithm AlphaGo, which has shown superhuman performance in zero-sum games. Our adaptation is capable of working in single-player and score based environments. In spite of the fact that AlphaGo versions are known to be computationally heavy, we manage to train the proposed framework with a single thread and GPU, while obtaining a solution that outperforms the state-of-the-art results in space utilization.
\end{abstract}

\section{INTRODUCTION}
\label{sec:intro}
The 3D Bin Packing Problem (3D-BPP) \cite{scheithauer1991three} is the 3D extension of the 1D-BPP and 2D-BPP versions (both versions widely studied). Bin Packing is one of the most demanded problems in the industry due to the wide applicability in logistics and manufacturing industries. Computationally, the problem is known to be strongly NP-Hard due to the huge search space. The problem mainly consist on packing a set of variable size 3D cuboid items in a sequence $\mathcal{I}$ into a finite bin, optimizing the space utilization, where W, L, H are the bin dimensions and $d_i=[w_i,l_i,h_i]^{\intercal}$ are the $i^{th}$ item dimensions. Typically, the size of the item is constrained to the bin size as follows; $w_i \leq W$, $l_i\leq L$ and $h_i\leq H$. For simplicity, the bin and items are discretized in voxels, so that $w_i, l_i, h_i, W, L, H \in Z^+$  are positive integers.

There are multiple variations of the problem definition. For instance, the problem can be formulated as a perfect information game where all items are known at any point in time; other settings may allow the relocation of items that have already been packed. In our work, we opt for a more practical definition based on \cite{zhao2020online}, where the decisions are irreversible and items are delivered in sequence one by one (online), such that we give special attention to the immediate items $\mathcal{B} \subset \mathcal{I}$. In practice, a conveyor belt carries the item sequence to a robotic arm located at the head of the line. This setup disallows the selection of items late in the sequence or having knowledge about the entire sequence. In the simple case scenario, only the most instant item in the sequence is considered $b = |\mathcal{B}| = 1$, we refer to this task as a single-item. 

\begin{figure*}[t]
\includegraphics[width=17cm]{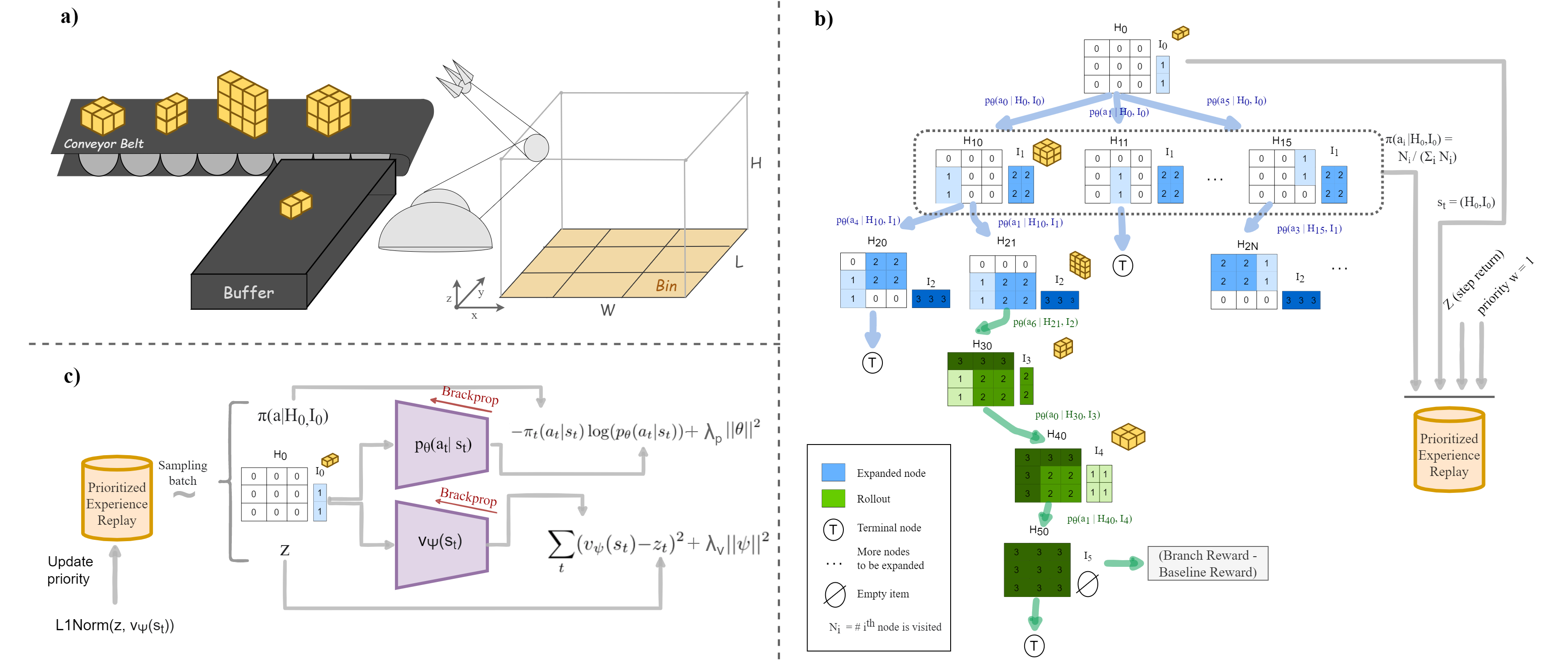}
\caption{Overall training example diagram of a $3 \times 3 \times 3$ bin. \textbf{a)} shows a simplified version of a conveyor belt with 4 items and buffer size of 1 item. \textbf{b)} depicts the state space exploration via MCTS for the first step with a single-item as node definition. A node is represented by the height map and the available items. Once the episode is terminated, the policies, states and returns of every step are stored in a prioritized replay storage. \textbf{c)} represents the update of the parameters $\theta$ and $\psi$ by sampling batches with priority from the experience storage.}
\label{fig:main}
\end{figure*}

A buffer located at the end of the conveyor belt can be placed to allow item selection of a small subset of immediate items, we define $b > 1$ as the multi-item selection task (Fig \ref{fig:main} a)). The buffer idea is inspired by human packers, as they can freely move around and select the most promising surrounding item. In addition, robotic arms are generally capable of rotating items ($z$ coordinate), we denominate $k$ as the number of reorientations, being $k=0$ when no reorientation is allowed. By enlarging both state representation and the action space, the algorithm can build more complex strategies, which can potentially lead to higher performance.

In this work, we adapt the well-known AlphaGo algorithm, to tackle various definitions of the online 3D-BPP with limited computation resources. In our adaptation, we focus on extending the algorithm to operate in single-player and score-based environments with known dynamic's model, rather than two player zero-sum games as it was originally defined. Likewise, we rely on Monte-Carlo Tree Search (MCTS) with rollouts to selectively explore the huge search space and provide approximations to the optimal policy. The experience collected by the algorithm is then augmented and inserted in a priority experience replay. Our experiments reveal that our adaptation surpasses the previous state-of-the-art results for single-item selection and item reorientation tasks across multiple datasets. Finally, when considering a buffer, the performance is observed to increase with respect to the single-item selection policy.

The \textbf{contributions} of our work are summarized as follows:
\begin{itemize}
    \item A new 3D-BPP framework based on a buffer to allow multi-item selection located at the end of the conveyor belt. The buffer benefits the agent in the following two ways: (1) building a more complex policy by expanding both action and state representations, (2) the episode can be continued even when one item does not fit in the current bin configuration.
    \item A model-agnostic data augmentation strategy that exploits the symmetries of the problem in order to improve in sampling efficiency and generalization.
    \item We empirically demonstrate that model-based methods, more concretely an adaptation of AlphaGo, can outperform the best so far presented model-free method in the bin packing problem with limited resources.
\end{itemize}

\section{BACKGROUND}
\noindent
\textbf{Traditional Bin Packing methods} have been extensively studied for solving 1D, 2D and 3D BPP. Due to the complexity of the problem (strongly NP-Hard), brute force algorithms are commonly discarded, instead heuristics models have been the principal option for many years. In particular to the 1D version, the algorithm must place items (represented by an integer) in available bins, however, if no feasible placement is found, a new bin must be initialized. Some popular algorithms are, Next-fit \cite{johnson1973near}, First-fit \cite{dosa2013first} and Best-fit \cite{dosa2014optimal}. The more generalized versions, 2D and 3D BPP, add extra complexity as the placement is constrained to the space and stability. The first heuristic 3D-BPP was presented in \cite{scheithauer1991three}, leading to the development of other successful algorithm like Tabu-Search \cite{crainic2009ts2pack}, guided local search \cite{faroe2003guided} and a combination of both Best and First fit in 3D \cite{dube2006optimizing}. Yet, machine learning algorithms are not the most popular option in the literature for many combinatorial problems.
\medbreak

\noindent
\textbf{Deep RL} is constituted by two entities, the agent and the environment, where the agent makes decisions within the environment, aiming to change the environment's state in such a way that the notion of cumulative reward is maximized throughout the agent's life (episode). More concretely, Deep RL consists of an agent  whose learning algorithm mainly relies on deep neural networks. RL algorithms are typically subdivided into two categories, i.e. model-free and model-based methods. The first has no assumptions about the environment's dynamics and as a consequence, the algorithm focuses solely on learning the optimal behavior or policy. model-free methods are further subdivided into value-based and policy optimization approaches. While value-based opts to estimate the value (discounted future reward) associated to a given state or state-action pair \cite{mnih2013playing}, policy optimization method explicitly learns the optimal state-action mapping \cite{wu2017scalable,schulman2017proximal}. In addition, a method can be categorized as on-policy or off-policy, where the main difference relies on whether the policy can learn from experience collected by other policies. This is the case of off-policy, which is considered more sampling efficient than on-policy. Regarding model-based methods, the algorithm can either assume the model of the environment is given \cite{silver2016mastering,silver2018general} or either can be learnt \cite{ha2018world,racaniere2017imagination}. The main advantage of model-based methods is that it allows the agent to plan ahead by virtually contemplating future scenarios. AlphaGo and later generalizations propose a deep RL model combined with MCTS for planning. These models have achieved super-human performance in board games where the state space is huge. Although the applicability is restricted to zero-sum games with two players, the most recent version MuZero \cite{schrittwieser2020mastering}, is able to master single-player games with unbounded and intermediate rewards in environments where the model is not given. In our work, we focus on a model-based approach with a given deterministic model. However, 3D-BPP has some fundamental differences from board games that we mention and address during this work. \medbreak

\noindent
\textbf{Deep RL for combinatorial optimization} has been recently motivated by the introduction of sequence-to-sequence models \cite{43155} and attention mechanisms. \cite{bello2016neural} proposed a framework to tackle NP combinatorial optimization problems, such as Travel Salesman Problem (TSP) or Knapsack with neural RL. Pointer Net was introduced by \cite{vinyals2015pointer}, a supervised model that leans on attention to solve TSP. Speaking of the research advancement in 3D-BPP with DRL is still in its early stages and to our knowledge, no model-based method has been attempted. \cite{hu2017solving}, inspired from previously mentioned works, presented a solution for the 3D-BPP relying on Pointer Net. Another relevant work is presented in \cite{zhao2020online}, the authors introduced an on-policy model-free DRL agent composed of an actor, a critic and a predictor, to predict action probabilities, value and feasibility mask respectively. Multiple on-policy and off-policy algorithms were trained, being the on-policy method \cite{wu2017scalable} the one with the highest performance. Not only they proposed a method but also they defined a realistic online 3D-BPP framework matching the specification of the real environment; due to this fact, we base our work on their framework. Furthermore, we make use of their datasets and method as the baseline for comparison in the experiment's section, since they claim state-of-the-art results.

\section{SIMULATORS}
The development of a realistic 3D Bin Packing simulator is somehow complex. The complexity resides primarily in the physics simulation, i.e. gravity. The bin distribution, the weight of each loaded item along with the gravity determines whether the packed items collapse. One way to detect instabilities a posteriori, i.e. once an item has already been packed, is by keeping track of undesired motion in the bin. This strategy can be utilized to effectively react to instabilities by penalizing via reward and early terminating the episode. Yet, the action mask is still required to avoid overlapping with walls and ceiling. An alternative strategy would be to design a preventive simulator, namely a simulator that assures a secure placement for each item by defining a more restrictive action mask according to a set of stability rules; as a result, no physics engine would be required.

\begin{figure}[t]
\includegraphics[width=8.5cm]{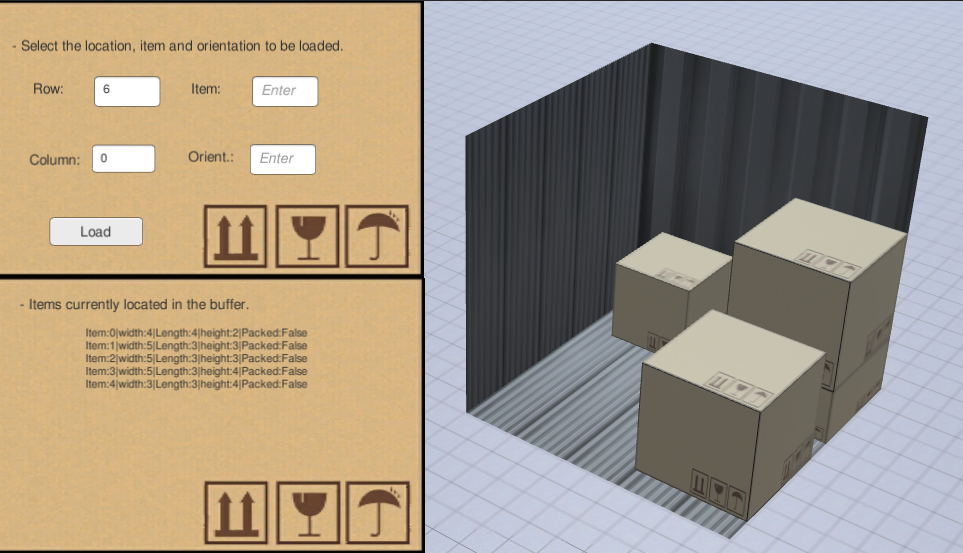}
\caption{Screenshot of the simulator UI. The left-hand side displays an interactive user panel, which is disabled during model training and inference. The bin and packed items are displayed on the right-hand side.}
\label{fig:env}
\end{figure}

We introduce two different simulators, a physics-based and a preventive-based simulator. The first is developed with Unity3D game engine and the machine learning agents (ML agents) extension \cite{juliani2018unity}. By default, Unity3D simulates gravity, collisions and even permits tracking object properties like the velocity. Another advantage of using Unity3D is that we can easily render the episode, which can be helpful for detecting bugs or abnormal agent behaviors during training and inference. The latter simulator is developed in Python. It consists of an multi-dimensional array representing bin distribution and the basic logic to carry out placements. Episode rendering is still not available for the preventive environment. Both simulators allow reorientation, item selection, lookahead, MCTS simulation and custom action mask definition. Finally, the sequences can be loaded from local files or generated on-the-fly.

\section{Datasets}
In real applications, items are delivered in random fashion and no assumptions about the sequence can be made, thus 100\% space utilization cannot be guaranteed. However, it would be convenient to know the maximum score for evaluation purposes. For this reason, \cite{zhao2020online} constructs three types of sequences, CUT-1, CUT-2 and Random Sequence (RS). CUT-1 and CUT-2 items are first generated via cutting-stock, that is to say, a bin sized cuboid  is randomly and recursively 'cut' until the sliced items match the size constraints. The difference between CUT-1 and CUT-2 relies on the sorting, while CUT-1 sorts the items by the Z coordinate in ascendant manner, CUT-2 sorts by stacking dependency: an item can be added if the supporting items are located earlier in the sequence. When two or more items have the same sorting priority, the next item is chosen at random. Both cutting-stock and sorting satisfy 100\% space utilization conditions. On the other hand, RS does not assume any conditions as the items are simply randomly generated, representing a more realistic dataset to train and test on.

All these three datasets consist of 2000 and 100 sequences for training and testing respectively. The item dimensions vary in the range between $[2,5]$ in all three dimensions, forming a set of 64 different items, while the bin resolution is $10 \times 10 \times 10$.

\section{METHOD}
In this section, we describe the AlphaGo adaptation, the first model-based approach applied to the 3D-BPP to our knowledge. The method is trained to solve three online tasks, i.e. single-item, item reorientation and multi-item selection with buffer. For simplicity, Fig. \ref{fig:main} shows a single-item training example of a sequence of 5 items. The figure is composed of three sections, a representation of the real environment, a single step of experience collection with MCTS and lastly, the policy update. To understand how Fig. \ref{fig:main} would look like with buffer and reorientation, the children of each node in the tree must be expanded with all new available actions (Cartesian product between locations, orientation and items) as well as adding multiple items in the state definition.

\subsection{Problem Setting}
More formally, the 3D-BPP is typically formulated as a Constrained Markov Decision Process (CMDP) due to the action constraints \cite{altman1999constrained}. A Markov decision process is often defined with the tuple $(\mathcal{S},\mathcal{A},\mathcal{R},\mathcal{P})$, where $\mathcal{S}$ is the set of states, $\mathcal{A}$ is the set of actions, $\mathcal{R}$ is the set of rewards and $\mathcal{P}$ is the model of the environment. $\mathcal{P}$ is the transition probability function $P(s'|s,a)$ where the next state $s'$ is conditioned to the current state $s$ and the action $a$. In our setup, we assume the transition model is given and is deterministic, meaning that a particular $(s,a)$ always leads to the same state $s'$.

To represent the bin state, we appeal to the height map representation $\mathcal{H}_t \in \mathbb{Z}^{W \times L}$, a matrix where each element indicates the height at the corresponding loading position (LP) in the bin. The action feature $a_t \in \mathbb{R}^{W \times L}$ provides with the placement probabilities, where each location represents the front-left-bottom of the input item (FLB). For example, given an item $i$ with dimensions $(w_i,l_i,h_i)$ and the LP $a_t=(x_t,y_t)$, the intended region covers the following area $h_t =(\mathcal{H}_{i,j})$  $[x_t \leq i < x_t+w_i,\:  y_t \leq j < y_t+l_i]\:$. The region $h_t$ is then updated $h_{t+1} = h_{max} + h_i$ where $h_{max}$ is the highest point in $h_t$. Also, the FLB must be readjusted when reorientation is applied, see Fig. \ref{fig:sym}. In addition, depending on the buffer configuration, we consider two different scenarios, the conventional online single item $b=1$ and the multiple item $b>1$ selection. For both cases and after the placement of each item, the next available item in the sequence is immediately added until the sequence is empty, in which case zero dimension item is included. For the sake of reducing item displacements, the next item takes the spot left by the last packed item, as a results, the robotic arm only requires one extra item displacement regardless of the buffer size.

The reward is in many cases the key to induce the agent to learn the correct behavior. In Atari games, the reward is matched with the game score. In particular to the BPP, the score is equivalent to the space utilization which is updated every time an item is placed. Therefore a non-zero reward is provided to the agent at every step. We reuse the reward definition of \cite{zhao2020online}, in which the reward is formulated to be proportional to the space utilization: $r_i = 10 \times V_i/V_B$, where $V_i$ is the volume of the $i^{th}$ item and $V_B$ is the volume of the bin. Once no available items or actions are found, a zero reward is provided and the episode is terminated.

At each time step, the environment provides the action mask, a binary mask responsible for ensuring a safer placement. In order to compare our results with \cite{zhao2020online}, we reuse their proposed set of feasibility rules. Nevertheless, these rules are not unique and also a physics-based simulator can relax the constraints as the instabilities can be detected a posteriori. The disjunctive rules for a particular FLB are as follows: 1) over 60\% of the area is supported including four corners, or 2) over 80\% of the area is supported including three or more corners, or 3) over 95\% of the item area is supported. Moreover, it is important to always make sure there is enough room w.r.t to the walls and ceiling of the bin.
 
\subsection{Neural Architecture}
We train a policy $p_\theta(a_t|s_t)$ with parameters $\theta$, built with a neural network architecture, to find the optimal mapping from states to actions. The input is represented as the feature combination of the height map $\mathcal{H}_t$ and the 3 dimensions per available item. Each item dimension is stretched out forming a plane with the same shape as $\mathcal{H}_t$. Finally, all features are stacked along the channel axis and normalized by the bin height H. If the remaining number of items in the buffer $\mathcal{B}$ is smaller than the buffer size $b$, the input feature is padded with zeros. The output tensor size is $W \times L \times b \times (k+1)$ covering every Cartesian combination of location, item and orientation. Once the input is forwarded through the policy net, the output  feature is element-wise multiplied with the action mask to remove infeasible LPs. Optionally, we train an independent value network $v_\psi(s_t)$ with parameters $\psi$, to predict the cumulative discounted reward $J(\pi) = E_{\tau \backsim \pi}\sum_{t=0}^{\infty} \gamma^t R(s_t,a_t)$, where $\gamma \rightarrow [0,1]$ is the discounted factor and $\tau$ is a state-action trajectory. Despite our method does not need the value function, it can be helpful combined with search algorithms during inference.

Both policy and value networks are constructed with a stack of 6 convolutional neural layers with 128 planes and $ReLU$ activation functions. The value network has a final linear layer with $tanh$ function to reduce and rescale the feature shape to a single dimension in $v_\psi(s_t) \in \mathbb{R} \rightarrow  [-1,1]$.

\subsection{Algorithm}
The method's training process is divided into two phases, the data generation (Fig.\ref{fig:main} b)) and the parameter update (Fig.\ref{fig:main} c)) phases. In the first phase, we consider the Monte-Carlo Tree Search strategy, which was recently introduced in \cite{silver2016mastering} for mastering the game of Go. The main goal of MCTS is to efficiently explore the huge search space looking for an approximation to the optimal policy $\pi$ for a given state. The search algorithm virtually executes a constant number of simulations from a starting state while keeping track of relevant statistics about the nodes, i.e. states. In contrast with the inference stage, during the training we assume MCTS is executed having perfect knowledge of the item sequence (known sequence), although the information accessible to $p_\theta$ and $v_\psi$ at each node is partial and it depends on the pre-defined buffer size. The expansion or selection of new nodes at each simulation is governed by the UCT Eq. \ref{eqn:uct}, composed by exploitation and exploration terms. While the exploitation term $Q(s,a)$ highlights the nodes with higher return, the exploration emphasizes nodes that have not often been visited $N(s,a)$. There are two main ways of evaluating a new expanded node in the tree, via rollout or via value network. Although value estimation has been proposed in later generalizations of AlphaGo, in our work we opt for rollout evaluation with action sampling, since we observed that rollouts are empirically more accurate than the value network for this problem (see subsection \ref{ssec:ablation}).

\begin{equation}
\label{eqn:uct}
U(s,a) = Q(s,a) + c_{puct} \, p_\theta(a|s) \frac{\sqrt{\sum_b N(s,b)}}{1 + N(s,a)}
\end{equation}

One fundamental difference between 3D-BPP and Go is that Go is a zero-sum game with two players, whereas 3D-BPP is score-based with intermediate rewards and single-player. In Go, the reward is given at the end of the episode, being $1$ for a win, $0$ for a draw and $-1$ for a loss. To convert the reward to a score style environment, First, before starting the MCTS simulations of a new sequence, the sequence is evaluated with a baseline policy. For simplicity, we reuse the policy $p_\theta(a_t|s_t)$, yet instead of sampling actions, the actions are deterministically selected according to the highest probability $arg\,max_{a_t} \,p_\theta(a_t|s_t)$. The return of a rollout is then computed by subtracting the baseline return from the rollout cumulative reward (from start to end of the episode). Intuitively speaking, the resulting score is positive if the explored trajectory ends up packing more items than the baseline and negative vice-versa. Rather than seeking trajectories that lead to 100\% utilization space (quite a hard task), we focus on finding a policy that incrementally improves the performance with respect to the baseline policy. Finally, once all simulations have been completed for a concrete step, the normalized visit count for every root action is calculated to generate the policy $\pi(a_t|s_t)$; in such a way that the more a node is visited, the higher is its probability.

A prioritized experience replay \cite{schaul2015prioritized} stores the experience collected in tuples $(s_t,\pi_t, z_t)$, where $s_t$ is the state, $\pi_t$ is the MCTS policies and $z_t$ is the rescaled episode ground truth discounted return $[-1,1]$. We set the $\gamma$ parameter to 1 in our experiments. Optionally, the samples can be augmented as described in Section \ref{ssec:aug}. Moreover, we can discard the episode samples depending on whether the MCTS policy outperforms the baseline score.

\begin{equation}
\label{eqn:loss}
\begin{split}
\mathcal{L} = \sum\limits_{t}(v_\psi(s_t) - z_t)^2 - \pi_t&(a_t|s_t) \log(  p_\theta(a_t|s_t)) \\
&+ \lambda_p||\theta||^2 + \lambda_v ||\psi||^2
\end{split}
\end{equation}

The second phase of the training proceeds right after a fixed number of data collection episodes. As depicted in Fig.\ref{fig:main} c), both $\psi$ and $\theta$ are optimized by minimizing the objective described in Eq. \ref{eqn:loss}, with batches sampled from the experience replay. Moreover, we set the learning-rate at the beginning of the training to $1 \cdot 10^{-3}$ for avoiding local minima and it is decreased as the training progresses.

\begin{figure}[t]
\includegraphics[width=8.5cm]{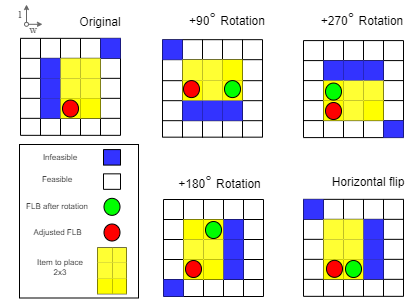}
\caption{A symmetry augmentation example of a 5x5 bin, placing a 3x2 item with 90º, 180º, 270º and horizontal flip augmentations. As depicted with the red and green circles, the FLB and the item orientation must be adjusted depending on the transformation applied. }
\label{fig:sym}
\end{figure}

\subsection{Data augmentation}
\label{ssec:aug}
Symmetries of the bin and items can be utilized to further expand the collected experience, similarly to Go or Othello \cite{silver2016mastering,silver2018general}. We propose rotation and flip transformations of the height map, items and action probabilities obtaining up to 8 times augmentation. The height map transformation and item features can be applied directly to the 2D matrix representation, however, the transformation of the action is not trivial. Go differs from 3D-BPP in that each stone occupies a single cell whereas items occupy a variable size. Fig \ref{fig:sym} shows an example of how a particular action FLB (marked in red) is shifted (marked in green) once a transformation is carried out. This is due to the fact that the FLB, which is the item's origin of coordinates, is relocated after the operation. This relocation depends on the item dimensions and the transformation. The following list summarizes all additional operations on the action probabilities and items, where the item dimensions are $d = [w,l,h]^{\intercal}$:

\begin{itemize}
\item The action probabilities after 90° rotation is readjusted by shifting left $l$ units. Also the item must be rotated w.r.t. the z coordinate.
\item The action probabilities after 180° rotation is readjusted by shifting left $w$ units and down $l$ units. The item does not need to be rotated since items are assumed to have 1 line of symmetry.
\item The action probabilities after 270° rotation is readjusted by shifting down $w$ units. The item must be rotated w.r.t. the z coordinate.
\item The action probabilities after a horizontal flip is readjusted by shifting left $w$ units and there is no need to rotate the item.
\end{itemize}

As a result of shifting the action feature left and/or down, the right and/or top empty cells are then set to zero to avoid overlap with the walls. Furthermore, the combination of flip and rotation transformations can be easily implemented by composition. In case the bin has a rectangular shape, i.e. $W \neq L$, 90º and 270º rotations may or may not be considered depending on whether the neural architecture can learn from different input tensor shapes.  


\begin{table}[t]
\caption{Simulator step delay}
\small
\begin{center}
\begin{tabular}{lll}
\textbf{}  &\textbf{Unity3DBP}  &\textbf{Py3DBP} \\
\hline \\
Single step         &$2.4 \pm 0.2$ ms  &$\textbf{0.13} \pm \textbf{0.02}$ ms \\
100 MCTS simulations  &$21.91\pm 13.1$ s  &$\textbf{0.80} \pm \textbf{0.51}$ s
\end{tabular}
\end{center}
\label{tab:sim}
\end{table}

\section{EXPERIMENTAL RESULTS}
We train the models with a single Intel i5 CPU at 3.70GHz $\times$ 6, 16 GiB of RAM and a single GPU Titan V. The training process takes about 1 day to reach the optimal policy for a single item task with 100 simulations per step. The training time complexity is sensitive to multiple factors, e.g. the number of masks to compute (1 per each item or orientation), the simulations per step and the duration of the episode.

Table \ref{tab:sim} compares the step delay for the mentioned simulators for two cases, a single step and 100 simulations. In spite of the fact that both environments reach \textbf{identical} performance under the same stability rules definition, it is straightforward to understand that the preventive simulator developed in Python is much faster than the physics-based, as it is much more simplistic, requiring no physics simulation nor rendering. The difference is negligible for a single step case, since the training bottleneck is the backward propagation phase rather than the experience collection phase. On the other hand, when utilizing MCTS, there is a drastic difference and the Unity3D simulator becomes impractical. For this reason, we use the preventive-environment for training our model-based approach. Nevertheless, the physics-based environment can still be useful, for example: simulating in inference mode, exploring new feasibility rules, training model-free models, training model-based combined with the preventive-based simulator for virtual steps and finally, simulating items with different weights. 

\begin{table}[t]
\small
\caption{Average \# packed items / Space util. (\%) comparison performance of single-item policies with and without reorientation for a $10\times10\times10$ bin.}
\begin{center}
\begin{tabular}{llll}

\textbf{k=0, b=1}  &\textbf{CUT-1}  &\textbf{CUT-2} &\textbf{RS} \\
\hline \\
Heuristics \cite{dube2006optimizing}    &15.15/59.8\%  &17.3/61.19\%  &\textbf{13.8/54.3}\%  \\
Model-free \cite{zhao2020online}     &19.1/73.4\%  &17.5/66.9\%  &12.2/50.5\%  \\
Ours     &\textbf{21.3/83.4}\%  &\textbf{18.0/69.9}\% &13.1/53.1\% \\
           &    &   & \\
\end{tabular}
\begin{tabular}{llll}
\textbf{k=1, b=1} &\textbf{}  &\textbf{} &\textbf{} \\
\hline \\
Heuristics \cite{dube2006optimizing}  &15.99/61.5\%  &17.62/62.5\%  &13.82/56.9\%  \\
Model-free  \cite{zhao2020online} &19.4/76.2\%  &18.1/70.2\%  &15.2/62.1\%  \\
Ours  &\textbf{22.1/85.6}\%  &\textbf{20.2/73.9}\% &\textbf{15.7/64.2}\%
\end{tabular}
\end{center}
\label{tab:perf}
\end{table}

\begin{table}[t]
\small
\caption{Average \# packed items / Space util. (\%) performance of our approach with multi-item selection: $k=0\,,b>1$.} \label{sim_tab}
\begin{center}
\begin{tabular}{llll}
\textbf{}  &\textbf{CUT-1}  &\textbf{CUT-2} &\textbf{RS} \\
\hline \\
b=1         &21.3/83.4\%  &18.0/69.9\% &13.1/53.1\% \\
b=2         &22.0/84.0\%  &20.2/71.5\% &14.6/57.6\% \\
b=3         &\textbf{22.5/85.7}\%   &\textbf{21.8/77.1}\%  &\textbf{15.8/62.1}\% \\
\end{tabular}
\label{tab:perf_buf}
\end{center}
\end{table}


\subsection{Single-item selection, $b=1$}
In the single-item task, i.e. buffer size $b=1$, only the next item in the sequence is known and can be selected. Once the item is placed, the next item in the sequence is considered. The episode terminates when the current item cannot be placed any longer, or the bin is perfectly packed. In addition, we allow reorientation in the z-coordinate of the item $k=1$, increasing the output size by a factor of two.

Table \ref{tab:perf} compares our model with the model-free baseline \cite{zhao2020online} and the heuristics method \cite{dube2006optimizing} for $k= 0$ and $k=1$ cases, across 3 datasets consisting of 100 episodes each. The performance is depicted with two metrics, the average number of packed items and the space utilization. Our method presents a superior performance for all datasets in both modalities. Our solution tends to pack from 0.5 up to 2.7 more packages on average than \cite{zhao2020online},  where the maximum difference is shown in CUT-1 with approximately 2 packages and 10\% of improvement for $k=0$ and $k=1$. Moreover, the performance of both approaches increases when the reorientation is allowed, specially in random sequence (RS), with more than 10\% improvement. In spite of the fact that the datasets CUT-1 and CUT-2 are designed to be perfectly packed without reorientation, the performance grows also in both datasets.

\subsection{Multi-item selection, $b>1$}
For buffer size $b>1$, the policy is trained to provide a placement probability for all buffered items. One of the advantages of including a buffer is that, even though the next available item in the sequence does not fit, the agent can still continue packing alternative items and potentially prepare the bin distribution to accommodate larger items. In practice, this reserved space cannot infinitely grow  due to the lack of accessible space around the robotic arm, as a result we only consider small buffer sizes $b=2$ and $b=3$.

Table \ref{tab:perf_buf} summarizes the performance of our trained policy for the cases $b=2, b=3$. The performance is observed to increase when the buffer size is expanded, being the policy $b=3$, the one with the best score for CUT-1, CUT-2 and RS. In particular to RS, the performance appears to be similar to reorientation $k=1,\,b=1$ case. Moreover, the buffer format and reorientation can be easily combined to develop more sophisticated policies by increasing the action space at the cost of more computational time complexity.

\begin{figure}[t]
\includegraphics[width=8.5cm]{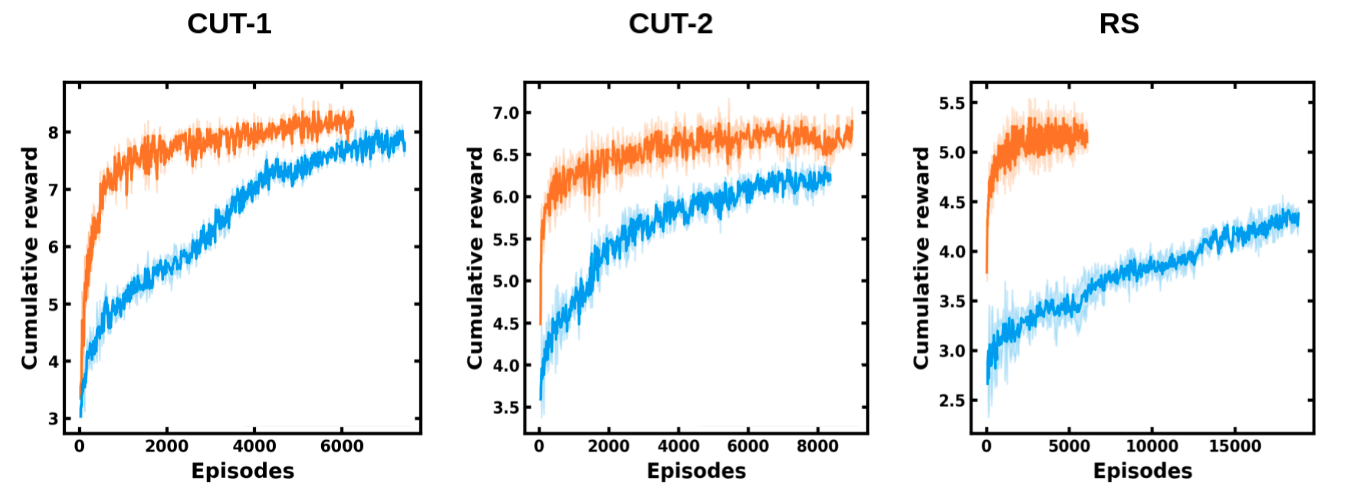}
\caption{Evolution of cumulative reward during the training progress for buffer size $b=1$ and all three datasets, where \color{orange}\textbf{--- }\color{black} is with and  \color{cyan}\textbf{--- }\color{black} is without data augmentation.}

\label{fig:training_prog}
\end{figure}

\begin{table}[t]
\small
\caption{Comparing multiple AlphaGo adaptations to the 3D-BPP.} \label{ablation}
\begin{center}
\begin{tabular}{llll}
\textbf{}  &\textbf{CUT-1}  &\textbf{CUT-2} &\textbf{RS} \\
\hline \\
Rollout with known sequences        &\textbf{83.4}\%  &\textbf{69.9}\% &\textbf{53.1}\% \\
Value with known sequences         &64.7\%   &65.9\%  &\textbf{53.1}\% \\
Rollout with stochastic sequences      &67.8\%   &66.0\%  &52.8\% \\
\end{tabular}
\label{tab:ablation}
\end{center}
\end{table}

\subsection{Ablation study}
\label{ssec:ablation}
Fig. \ref{fig:training_prog} depicts the training evolution of the cumulative reward per episode for two cases: with and without data augmentation. It is easy to notice that data augmentation speeds up the training for all three datasets. Not only accelerates but more importantly, achieves a higher performance for all cases and more notably when RS dataset is selected.

Table \ref{tab:ablation} compares the performance of 3 different AlphaGo variations in the 3D-BPP. As mentioned earlier, rollout estimation leads to better performance than value estimation yet with a little more overhead during training. One potential explanation could be the fact that the value network cannot accurately predict the future discounted reward with limited state definition. Finally, since the complete sequence is unknown during inference, we attempt to train the algorithm assuming stochastic item arrival modeled with pre-trained sequence models for CUT-1 and CUT-2 and uniform sampling for RS. The results show that using the actual sequence during training excels the stochastic sequence approach. There are two main reasons why we think stochastic sequences do not work well: first, a small perturbation or change in the sequence can drastically change the value estimation for a particular state. This brings us to the second issue, in order to somehow approximate the optimal policy under a very noisy environment, a very large number of simulations would be required (too expensive for our setup). To solve these issues, we simply use the corresponding sequence during training (not inference) despite adding little bias. We observe that the algorithm is capable of generalizing well to the unseen sequences

\section{DISCUSSION}
The proposed method has shown improvement in performance in three different datasets by slightly increasing the training time with limited resources. The algorithm learns the optimal policy with fewer simulations per step in comparison with AlphaZero for the given problem. Although the scalability issue of the proposed method for much higher bin resolutions is best left for future works, we may consider a potential way to tackle the issue, besides a simple scale-up of the hardware resources. That is, we can adopt the strategy proposed for the game of Go and Reversi in \cite{ben2021train}, in which the 2D board is represented into a graph structure and trained by a GNN-style model with the AlphaZero algorithm, such that the board size can be tractable and extended during inference while keeping the performance. 

The previous point brings us to the next point; there is a lack of standardized benchmarks for the 3D Bin Packing Problem. In order to compare our work, we selected a publicly available dataset \cite{zhao2020online}, however, we believe it would be necessary to build open-source datasets similarly to the computer vision domain. These tasks can be divided into multiple categories according to, for instance, bin resolutions, item ratios and modalities such as multi-bin, buffer and so on.

\section{CONCLUSIONS \& FUTURE WORK}
In this work we have proposed a new 3D bin packing framework enhanced with an item buffer. We have shown that by adding the buffer the agent reaches considerably higher packing performance while remaining compatible with real applications. In addition, we introduced a data augmentation strategy to improve the sampling efficiency and training speedup, benefiting setups with limited hardware resources. Also, we have demonstrated that the proposed AlphaGo adaptation is more accurate than model-free approaches for several online tasks. In discussion, the case of $k=0, b=1$ in Table \ref{tab:perf} shows the effectiveness of the proposed data augmentation and the model-based rollout with adapted AlphGo in improving performance, especially, for the case with a lesser degree of variability in sequence such as CUT-1. On the other hand, Table \ref{tab:perf_buf} and the $k=1, b=1$ case of Table \ref{tab:perf} indicate that both buffer and reorientation are also effective for improving performance. It is interesting to observe that the buffer and reorientation strategy tend to impact more on the case with a larger degree of variability in sequence such as CUT-2 and RS. We conjecture that provision of a higher degree of freedom in action selection incurs a more positive impact on the performance of RL, especially for dealing with larger environmental variations. 

For the future work, we aim to increase the computational power as well as testing our implementation in real industrial environments with larger bin sizes, wider variety of item sizes and higher space resolution in order to explore the full potential of the algorithm. As mentioned in the discussion section, we aim to propose standardized benchmarks for the 3D-BPP to more fairly compare algorithms for multiple sub-tasks and settings. Finally, we would like to adapt the proposed framework to a continuous action space setting.

\addtolength{\textheight}{-12cm}   

\bibliographystyle{IEEEtran}
\bibliography{references}

\end{document}